# DEEP FEW-SHOT LEARNING FOR BI-TEMPORAL BUILDING CHANGE DETECTION

M. Khoshboresh-Masouleh*, R. Shah-Hosseini

School of Surveying and Geospatial Engineering, College of Engineering, University of Tehran, Tehran, Iran – (m.khoshboresh, rshahosseini)@ut.ac.ir

**ABSTRACT:**

In real-world applications (e.g., change detection), annotating images is very expensive. To build effective deep learning models in these applications, deep few-shot learning methods have been developed and prove to be a robust approach in small training data. The analysis of building change detection from high spatial resolution remote sensing observations is important research in photogrammetry, computer vision, and remote sensing nowadays, which can be widely used in a variety of real-world applications, such as map updating. As manual high resolution image interpretation is expensive and time-consuming, building change detection methods are of high interest. The interest in developing building change detection approaches from optical remote sensing images is rapidly increasing due to larger coverages, and lower costs of optical images. In this study, we focus on building change detection analysis on a small set of building change from different regions that sit in several cities. In this paper, a new deep few-shot learning method is proposed for building change detection using Monte Carlo dropout and remote sensing observations. The setup is based on a small dataset, including bitemporal optical images labeled for building change detection.

## 1. INTRODUCTION

Current urban and rural space growth rates and obvious effects of building construction on different applications, such as building damage assessments, building change detection, population estimation, and urban planning continuous monitoring of building footprints are becoming ever more significant [1]–[6]. The interest in developing building change detection methods from bi-temporal optical images with different platform (e.g., UAV, Airborne, and Spaceborne) is rapidly increasing due to larger land-covers, and lower costs of remote sensing data [7]–[9]. Despite rapid advances in computer vision and remote sensing, change detection is a major challenge in mapping. However, building change detection from an optical remote sensing image with high accuracy and precision needs considerable efforts in developing robust methods. Fig. 1 shows different challenges of building change detection from optical images such as small change (cf. Fig. 1a), complex roofs (cf. Fig. 1b), multiscale change (cf. Fig.1c), and no-change (cf. Fig. 1d).

Owing to the advances of deep learning and the new datasets released in computer vison, remote sensing and photogrammetry, new insights have been presented in the field of building change detection for bi-temporal data [10]–[14]. Although some efforts have been devoted to the development of deep learning approaches, little attention has been devoted to deep few-shot learning for change detection.

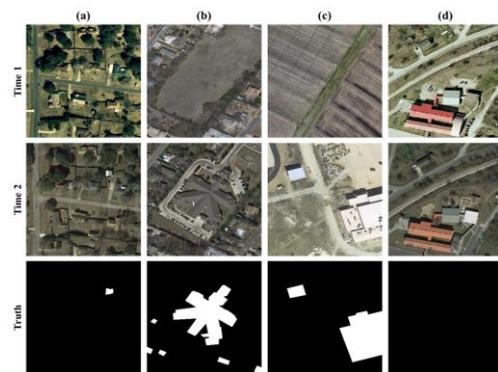

Figure 1. Example bi-temporal data from the LEVIR-CD dataset for building change detection.

Table 1 presents the overview of the methods recently published in applied convolutional neural networks for building change detection from bi-temporal images, based on highlighting characteristics and precision with focus on research contributions. Although the related approaches are robust, but it's not still a good performance for building change detection, particularly, in robust building change detection from a small training dataset.

| Method | Precision | Reference |
|---|---|---|
| Double CNN | 50% | [15] |
| FC-EF | 62% | [16] |
| FC-Siam-conc | 52% | [16] |
| FC-Siam-diff | 58% | [16] |
| Multiscale CNN | 63% | [17] |

Table 1. Overview of recent building change detection methods using bi-temporal RGB images

Building change detection approaches for bi-temporal data need to be analyzed in different regions with uncertainty estimation. Therefore, despite the previous approaches that try to extract building change without uncertainty map from small regions with deep learning models and large-scale dataset, in this paper, we focus on building change detection analysis on a small set of building change from different areas that sit in several cities. In this study, to build generalized and effective building change detection models in remote sensing, deep few-shot learning methods have been developed and prove to be a robust method in small training data.

---

* Corresponding author

## 2. FEW-SHOT CNN

In real-world applications such as building change detection, annotating images is very expensive. This issue becomes more important in the tasks of pixel-wise classification, where dense labels are harder to annotate. The goal of few-shot learning in building change detection is to predict a binary map of a change class given a few pairs of support and query images containing the same change class and the binary truth maps for the support images [18]. Existing few-shot approaches generally learn from a handful of support images and then feed learned information into a parametric module for segmenting the query [19].

To build generalized deep learning models in building change detection, deep few-shot learning models have been developed and prove to be a robust tool in small labeled data. One efficient method is to fine-tune the pre-trained network. In this study, we use the MultiScale-Net [3] as a specialized backbone to take advantage of its strong capacity for building detection. Unlike the previous methods that use general CNN series as the backbone, we use the specialized model for building detection as the backbone.

Fig. 2 illustrates an overview of the Few-Shot CNN for building change detection. There are two same encoder block tied convolutional blocks for extracting features from $t_1$ and $t_2$ images. The proposed method learns the representation from image pair, and then tries to find the relationship between them. Using the residual blocks in convolutional neural networks, despite the improvement of accuracy, significantly increases the cost of calculations. The increased cost of calculations significantly affects the integration of the residual blocks with the convolutional blocks. Assuming that the convolutional block includes two convolutional filters, the input value passes through two filters and, then, is added to the initial value. As the theory suggests, the processed value is added to the initial value merely for preventing the reduction of the features created by the convolutional filters. This process makes it difficult to employ these layers in encoding-decoding networks. In order to promote the use of residual blocks in encoding-decoding networks, a new method called convolutional residual blocks based on depth dropout method was used in this study [20].

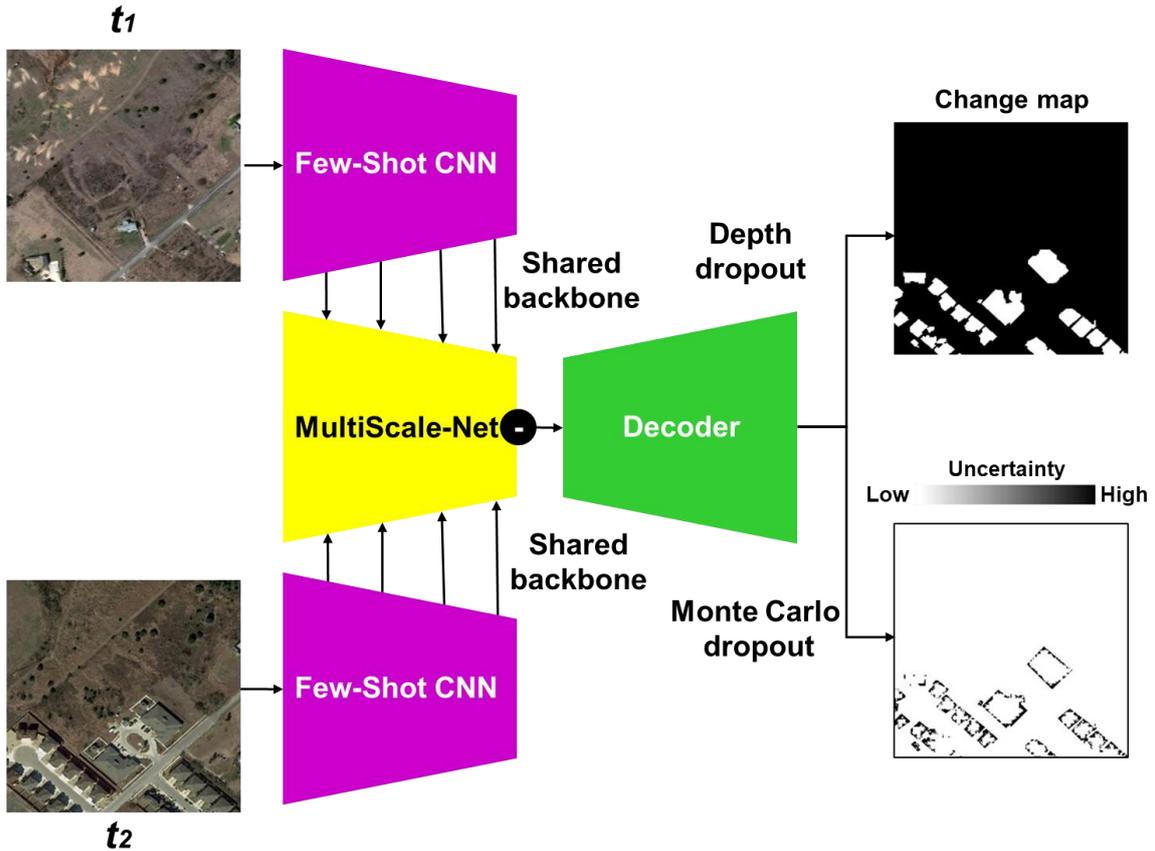

Figure 2. Few-Shot CNN. This example shows the performance of the Few-Shot CNN on the LEVIR-CD. The network takes bi-temporal RGB images as input, and outputs a building change map

## 3. UNCERTAINTY ESTIMATION

Uncertainty estimation is a crucial step in the evaluation of the robustness for deep and few-shot learning models in remote sensing, especially when applied in risk-sensitive areas, such as building change detection [8], [21], [22]. Knowing the confidence with which we can trust the building change detection is important for decision making in remote sensing [23]. There are some approaches to uncertainty estimation for deep learning proposed, but most of them need to sample several times, which is destructive to bi-temporal applications [24]. In this paper, we focus on Monte Carlo dropout [25] for uncertainty estimation in building change detection.

In this study, we use the Monte Carlo dropout as the aleatoric and epistemic uncertainty estimator for building change detection. In the image segmentation, aleatoric is an uncertainty metric of the intrinsic, irreducible noise found in the image, usually associated with the image acquisition process. Epistemic is uncertainty over the actual values of a model's parameters arising from the finite size of the training datasets [26]. As depicted in Fig. 2, total uncertainty obtained by the Monte Carlo dropout.

## 4. RESULTS AND DISCUSSION

To evaluate the Few-Shot CNN, forty test images were selected from study areas. The features of these test images include small change, complex roofs, multiscale change, and no-change.
To improve the training process of MultiScale-Net model, pretrained weights have been used in [3]. Pretrained weights have been trained on the basis of MultiScale-Net model training with 2868 patches of very high-spatial resolution images for building detection from different optical sensors. Table 2 shows the splitting statistics of the dataset.
Few-Shot CNN was trained with ADAM optimizer [27] using the default parameters and with a batch size of 64 for 250 epochs for building change detection. The generated dataset includes 190 bi-temporal RGB images that are from the LEVIR-CD [10], with a size of $512 \times 512$ pixels and a spatial resolution from 0.5 m that covers different cities in the United States.

As shown in Fig. 3, four samples are selected from the test area for assessing the performance of the proposed method. The evaluation metric of intersection over union (IoU) (for change map) and entropy (for uncertainty map) are used to evaluate the performance of the proposed method. The average IoU and entropy for all scenes are about 92.4% and 0.12, respectively. The results using the LEVIR-CD dataset demonstrate the more reasonable accuracy achievement of the Few-Shot CNN to building change detection with small training data.

| Dataset | Reference | Training | Test |
|---|---|---|---|
| IND | [8] | 300* | - |
| WHU-I | [28] | 204* | - |
| Inria | [29] | 200* | - |
| LEVIR-CD | [10] | 150 | 40 |

Table 2. Dataset splitting statistics. * Denotes used only for the pretrained step

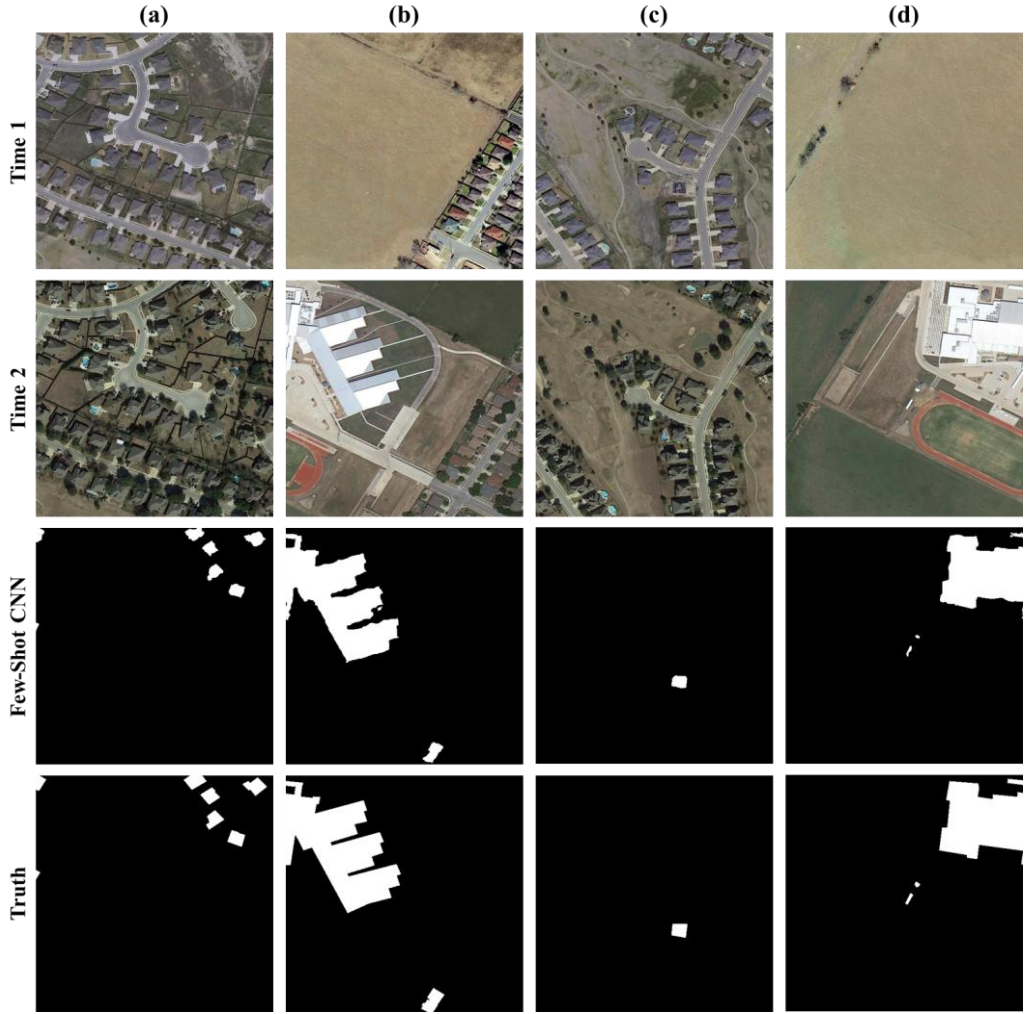

Figure 3. Building change detection results of the LEVIR-CD in four samples

## 5. CONCLUSION

In this study, using the fusion of Few-Shot CNN, Depth dropout, and Monte Carlo dropout, a novel method was presented and implemented for building change detection from bi-temporal RGB images. Then, the proposed network was tested on four remote sensing data sets with high spatial resolution and different challenges in mapping. The purpose of this study is to investigate the capabilities of this algorithm in the field of change detection. This is important because the access of some countries to high spatial resolution remote sensing image is far better and easier than that of other countries, and this approach can make the automatic mapping process possible with a relatively inexpensive remote sensing data source, with full coverage of different regions.